\pdfoutput=1

\documentclass[11pt]{article}

\usepackage{EACL2023}

\usepackage{times}
\usepackage{latexsym}

\usepackage[utf8]{inputenc}

\usepackage{microtype}

\usepackage{inconsolata}

\usepackage{amsmath,amsfonts, amssymb}
\usepackage{soul}
\usepackage{filecontents}
\usepackage[inline]{enumitem}
\usepackage{pgfplots, pgfplotstable}
\usepgfplotslibrary{groupplots}

\usepackage[ukrainian,english]{babel}
\usepackage{algorithmic}
\usepackage{algorithm}
\usepackage{array}

\newcommand{\specialcell}[2][c]{
  \begin{tabular}[#1]{@{}c@{}}#2\end{tabular}}

\title{RedPenNet for Grammatical Error Correction: Outputs to Tokens, Attentions to Spans}

\author{Bohdan Didenko \\
  WebSpellChecker LLC / Ukraine \\
  \texttt{bogdan@webspellchecker.net} \\\And
  Andrii Sameliuk \\
  WebSpellChecker LLC / Ukraine \\
  \texttt{andrii.sameliuk@webspellchecker.net} \\}

\begin{document}
\maketitle
\begin{abstract}
The text editing tasks, including sentence fusion, sentence splitting and rephrasing, text simplification, and Grammatical Error Correction (GEC), share a common trait of dealing with highly similar input and output sequences. This area of research lies at the intersection of two well-established fields: (i) fully autoregressive sequence-to-sequence approaches commonly used in tasks like Neural Machine Translation (NMT) and (ii) sequence tagging techniques commonly used to address tasks such as Part-of-speech tagging, Named-entity recognition (NER), and similar. In the pursuit of a balanced architecture, researchers have come up with numerous imaginative and unconventional solutions, which we're discussing in the Related Works~\ref{related} section. Our approach to addressing text editing tasks is called RedPenNet and is aimed at reducing architectural and parametric redundancies presented in specific Sequence-To-Edits models, preserving their semi-autoregressive advantages.
Our models achieve $F_{0.5}$ scores of 77.60 on the BEA-2019 (test), which can be considered as state-of-the-art the only exception for system combination~\citep{qorib-etal-2022-frustratingly} and 67.71 on the UAGEC+Fluency (test) benchmarks.

This research is being conducted in the context of the UNLP 2023 workshop, where it was presented as a paper as a paper for the Shared Task in Grammatical Error Correction (GEC) for Ukrainian. This study aims to apply the RedPenNet approach to address the GEC problem in the Ukrainian language.
Public data related to this article may appear over time in this GitHub repository~\footnote{\url{https://github.com/WebSpellChecker/unlp-2023-shared-task}}.
\end{abstract}

\section{Introduction}
The GEC challenge has been tackled with various techniques, including the traditional Autoregressive (AR) Neural Machine Translation (NMT) using the transformer architecture~\citep{https://doi.org/10.48550/arxiv.1706.03762}, as well as additional methods that we refer to collectively as Inference Optimized (IO). The existing IO methods for GEC can be broadly categorized into two groups, as described further.

The first group is non-autoregressive Feed Forward (FF) approaches which involve a single forward pass — through the model and provides token-level edit operations, such as the approach proposed in~\citep{DBLP:journals/corr/abs-1910-02893},~\citep{https://doi.org/10.48550/arxiv.2005.12592}. The advantage of FF approaches is their fast inference speed. However, their limitations lie in how they maintain consistency between interrelated edits, which leads to the need for iterative sentence correction approaches. The iterative sentence correction process solves some issues with interrelated corrections. However, it introduces new challenges. The absence of information about the initial input state could potentially lead to substantial modifications of the text meaning and structure, including rewording, word rearrangement, and the addition or removal of sentence components.

The second category consists of Inference Optimized Autoregressive (IOAR) models, which can be further separated into two subcategories: (i) sequence-to-edits (SeqToEdits). This category encompasses works such as~\citep{https://doi.org/10.48550/arxiv.1909.01187},~\citep{chen-etal-2020-improving-efficiency},~\citep{stahlberg-kumar-2020-seq2edits}, and the RedPenNet model examined in this paper; (ii) the recently proposed Input-guided Aggressive Decoding (IGAD) approach~\citep{https://doi.org/10.48550/arxiv.2205.10350}, which has been proven effective for GEC tasks, as demonstrated in the study~\citep{https://doi.org/10.48550/arxiv.2106.04970}. More information about these model categories can be found in the Related Work~\ref{related} section.

In our study, we propose RedPenNet, which is an IOAR model of the SeqToEdits subtype. RedPenNet utilizes a single shallow decoder~\citep{https://doi.org/10.48550/arxiv.2006.10369} for generating both replacement tokens and spans. During the generation of edit tokens, the encoder-decoder attention weights are used to determine the edit spans. For these attentions, pre-softmax logits are fed as inputs to a linear transformation which predicts the position of the edit in the source sentence. This approach is similar to the method described in Pointer Networks~\citep{https://doi.org/10.48550/arxiv.1506.03134}. Additionally, we train compact task-specific decoder BPE vocabularies to reduce the cost of the pre-softmax dot operation, making it more efficient for predicting replacement tokens. The RedPenNet model is also capable of tackling the challenge of Multilingual GEC ~\citep{rothe-etal-2021-simple}. To achieve this, specialized shallow decoders need to be trained for different languages. This gives the ability to use a single model with a multilingual pre-trained encoder and language-specific decoders.

Our proposed solution has a design that enables converting the input sequence into any output sequence, achieving competitive results in solving the GEC task.

\section{RedPenNet}
\subsection{General} \label{general}
Instead of predicting the target sequence directly, the RedPenNet model generates a sequence of $N$ 2-tuples $(t_n, s_n)\in V\times \mathbb{N}_0$ where $t_n$ is a BPE token obtained from the pre-computed decoder vocabulary~\ref{vocabularies} and $s_n$ denotes the span positions. In the RedPenNet approach, we define each of the $J$ edit operations $\mathbf{e}$ as a sequence of $C$ 2-tuples $(t_c, s_c)$, where $2 \leq C \leq N$. The first token for each edit $e_j$ is represented as $t_{c=0} = \mathtt{SEP}$. As previously mentioned, in our approach, a single edit can consist of multiple tokens. However, to determine the span of a single edit, only two positions are required: $s_{start}$ and $s_{end}$. To accomplish this, we impose the following constraints for each $e_j$: $s\_{start} = s_{c_0}$ and $s\_{end} = s_{c_1}$. The remaining $s_c$ values for $c \in {2, ..., C}$ are not considered. In RedPenNet, if a correction requires inserting text at a position $n$ in the source sequence, it is expressed as $s_{start} = s_{end}$. To handle the deletion operation, a special token {DEL} $\in V$ is used, which is equivalent to replacing the span with an empty string. If the input text is error-free, RedPenNet generates an {EOS} token in the first AR step, thereby avoiding unnecessary calculations. 

The iterative process of applying edits to the source sequence is illustrated in Algorithm~\ref{alg:applEdits}. Also, the process of generating GEC edits using the RedPenNet architecture can be visualized with the help of the following illustration~\ref{tab:decoding}.

In the case of RedPenNet, similar to the {S}eq2{E}dits approach~\citep{stahlberg-kumar-2020-seq2edits}, it is important to maintain a monotonic, left-to-right order of spans and ensure that {SEP} tokens are never adjacent to each other and the final edit token is always {EOS}. None of our models generated invalid sequences during inference without any constraints, as it is also the case with {S}eq2{E}dits.

\begin{algorithm}[t!]
\caption{editsToCorrect()}
\label{alg:applEdits}
\begin{algorithmic}[1]
\STATE{$s\_start\gets 0$}
\STATE{$s\_end\gets 0$} \COMMENT{ Initialize spans}
\STATE{$\mathbf{y}\gets\mathbf{x}$} \COMMENT{Initialize $\mathbf{y}$ as tokenized input}
\STATE{$\mathbf{z}\gets\epsilon$} \COMMENT{Initialize $\mathbf{z}$ edit seq with the empty string.}
\FOR{$n\gets 1$ \TO $N$}
    \IF {$t_n = \mathtt{EOS}$}
        \RETURN{$\mathbf{y}$}
    \ELSIF{$t_n = \mathtt{SEP}$}        
        \STATE{$\mathbf{y}_{s\_start}^{s\_end}\gets\mathbf{z}$}
        \STATE{$\mathbf{z}\gets\epsilon$}        
        \STATE{$s\_start\gets{s_n}$}
        
    \ELSE 
        \IF{$t_n \neq \mathtt{DEL}$}
            \STATE{$\mathbf{z}\gets\texttt{concat}(\mathbf{z}, t_n)$}
        \ENDIF
        \IF{$t_{n-1} = \mathtt{SEP}$}
            \STATE{$s\_end\gets{s_n}$}
        \ENDIF
    \ENDIF
\ENDFOR

\end{algorithmic}
\end{algorithm}

\begin{table*}[!ht]
    \centering
    \small  
    \label{tab:sentence}
    \renewcommand{\arraystretch}{1.2} 
    \begin{tabular}{|c|*{18}{@{\hspace{0.1cm}}c@{\hspace{0.1cm}}|}{c|}}
        \hline
        Positions & 1 & 2 & 3 & 4 & 5 & 5 & 6 & 7 & 8 & 9 & 10 & 11 & 12 & 13 & 14 & 15 & 16 & 17 \\ \hline
        Input Sentence & \st{In} & \st{a} & other & hand & $\varnothing$ & many & \st{of} & stars & sold & their & privacy & to & earn & more & and & more & money & . \\ \hline        
    \end{tabular}
    \caption{Visual representation of the tokenized encoder input sequence that includes visual markings, intended to improve the clarity of the editing process.}
    \vspace{0.2cm}

    \small
    \label{tab:decoding}
    \begin{tabular}{|l|l|l|l|l|l|l|l|l|}
        \hline
        Autoregressive Steps & 1 & 2 & 3 & 4 & 5 & 6 & 7 & 8\\ \hline
        
        Input Tokens & \tiny{SOS} & \tiny{SEP} & On & the & \tiny{SEP} & , & \tiny{SEP} & \tiny{DEL} \\ \hline        
        Input Spans & 0 & 1 & 3 & $\varnothing$ & 5 & 5 & 6 & 7 \\ \hline
        Output Tokens & \tiny{SEP} & On & the & \tiny{SEP} & , & \tiny{SEP} & \tiny{DEL} & \tiny{EOS} \\ \hline
        Output Spans & 1 & 3 & $\varnothing$ & 5 & 5 & 6 & 7 & $\varnothing$ \\ \hline        
    \end{tabular}
    \caption{This example depicts a step-by-step demonstration of a RedPenNet autoregressive inference that encompasses multi-token edits, insertions and deletions.}
\end{table*}

\subsection{Encoder}
The utilization of pre-trained language models has been consistently shown to improve performance on a range of NLP downstream tasks, including GEC, as observed in numerous studies. To train RedPenNet, we deployed pre-trained models from the HuggingFace transformers library\footnote{\url{https://huggingface.co/models}}. We have observed that models trained on Masked Language Modeling (MLM) tasks perform the best as encoders for the RedPenNet architecture. Therefore, in this work, we focus solely on this family of models. The availability of a range of models within the HuggingFace library offers the flexibility to choose a pre-trained model based on the required size, language, or a multilingual group. This opens up the potential for RedPenNet to (i) create multilingual GEC solutions using language-specific decoders with a single multilingual encoder, and (ii) construct RedPenNet model ensembles based on different pre-trained models with comparatively less efforts.

\subsection{Decoder} \label{decoder}
A transformer decoder stack was used as for the decoder in the RedPenNet model. During the training phase, the model learned an autoregressive scoring function $P(\mathbf{t}, \mathbf{s} | \mathbf{x}; \mathbf{\Phi})$, which is implemented as follows:
\begin{equation*}
\begin{split}
    &\mathbf{\Phi\ast} = \mathrm{arg}\text{ }\underset{\mathbf{\Phi\ast}}{\mathrm{max}}\text{ }\mathrm{log}P(\mathbf{t}, \mathbf{s} | \mathbf{x}; \mathbf{\Phi}) \\ = &\mathrm{arg}\text{ }\underset{\mathbf{\Phi\ast}}{\mathrm{max}}\sum_{n=1}^N\mathrm{log}P(t_n, s_n | t_1^{n-1},s_1^{n-1},\mathbf{x}; \mathbf{\Phi}) 
\end{split}
\end{equation*}

where $\mathbf{t} = (t_1, . . . , t_n)$ represents the sequence of ground-truth edit tokens with {SEP} tokens that are used to mark the start of each edit. Additionally, $\mathbf{s} = (s_1, . . . , s_n)$ indicates the sequence of ground-truth span positions, which denote specific ranges in the input sequence $\mathbf{x}$.

In line with the standard Transformer architecture, the previous time step predictions are fed back into the Transformer decoder. 
At each step $n$, the feedback loop consists of the BPE token embedding of $t_{n-1}$, which is combined with a decoder-specific trainable positional encoding embedding $p_{n-1}$. The resulting sum is then concatenated with the span embedding of $s_{n-1}$.

The span embedding on step $n$ can be defined as:
\begin{equation*}
\begin{split}
 &\mathbf{s\_emb}_n = s\_mask_{n} \cdot \mathbf{x\_emb}_{s_{n-1}} \\
&s\_mask_{n} =
\begin{cases}
0, & \text{if } n = 0\\
1, & \text{if } 0 < n \leq 2\\
0^{|tid_{n-1}-b|} \\ +0^{|tid_{n-2}-b|}, & \text{otherwise}
\end{cases}
\end{split}
\end{equation*}

where $tid_n$ is an index of token $t_n$ in decoder vocabulary $V$ and is denoted as $\mathbf{tid} = \mathrm{index}(\mathbf{t}, V)$, $b$ is an index of {SEP} token in $V$ and $\mathbf{x\_emb}_{s_{n-1}}$ is a vector embedding corresponding to $x_{s_{n-1}}$ token.
 In other words, there are two cases in how span embedding takes values depending on the preceding token sequence: (i) by using the embedding of the token $x_{s_{n-1}}$ from the encoder input sequence when $t_{n-2} = SEP \lor t_{n-1} = SEP$ or (ii) by using a zero-filled embedding $\epsilon$ with the same dimensions $D$ as $x_n$. During training, similar to the MLM task, a binary spans target mask for spans sequences are used to regulate the given logic.
 
 As a result, the inputs of the decoder at step $n-1$ can be expressed as follows: 
 \begin{equation*}
\begin{split}
&\text{Concat}(\mathbf{t\_emb}_{n-1} + \mathbf{p}_{n-1}, \mathbf{s\_emb}_{n-1}) \\ = 
&[t\_emb_{n-1,1} + p_{n-1,1}, ..., t\_emb_{n-1,D} + p_{n-1,D}, \\ 
&{s\_emb}_{n-1,1}, ...,{s\_emb}_{n-1,D}] \in \mathbb{R}^{2D}
\end{split}
\end{equation*}
 
The technique of utilizing the pre-softmax attention weights from an encoder-decoder attention layer to represent the probabilities of positions in the input sequence was introduced in Pointer Networks~\citep{https://doi.org/10.48550/arxiv.1506.03134} and later applied to the GEC task in the {S}eq2{E}dits approach~\citep{stahlberg-kumar-2020-seq2edits}. Additionally, to increase the number of trainable parameters at this stage, a dense layer has been added to the bottom of the spans output linear transform.

\section{Training Decoder BPE Vocabularies} \label{vocabularies}
In the traditional implementation of the Transformer model, a shared source-target vocabulary is utilized for both the decoder and the encoder, as described in~\citep{https://doi.org/10.48550/arxiv.1706.03762}. It is evident that the pre-softmax linear transformation required to transform the decoder output into predicted next-token probabilities is computationally expensive. Its computational complexity can be expressed as $O(d \cdot v)$, where $d$ is the output dimension of the model and $v$ is the decoder vocabulary size.

If the GEC task is approached by generating correction strings for the edits and using autoregressive decoding for this purpose, we tend to think that the information entropy of the generated sequences will be significantly lower compared to that of the input sequences. Our belief is based on the following two assumptions:
\begin{enumerate}
\item People tend to make mistakes in similar phrases and words.
\item Corrected versions of spelling words are statistically more frequent and can be represented by fewer BPE tokens.
\end{enumerate}
Therefore, a smaller BPE vocabulary will be sufficient to create efficient representations of sequences of corrections. In section~\ref{ukr_gec_vocabularies}, we test this hypothesis on one of the languages, as the example shows.

\section{Related Work} \label{related}

In the context of the GEC task, the closest family of approaches to RedPenNet is the Autoregressive approaches, specifically the SeqToEdits subtype. They include models such as Lasertagger, Erroneous Span Detection and Correction (ESD\&ESC), and Seq2Edits comparison with which is important for understanding the impact of our work. These models share the advantage that the number of autoregressive steps are based on the number of necessary edits to the original text, rather than the length of the input text. We will evaluate each of these approaches to the GEC problem in this section. In this section, we will also discuss the Aggressive Decoding approach, which has evolved from the traditional sequence-to-sequence approach.

The {S}eq2{E}dits~\citep{stahlberg-kumar-2020-seq2edits} approach predicts a sequence of N edit operations autoregressively from left to the right. Each edit operation is represented as a 3-tuple (tag, span, token) that specifies the action of replacing. The approach allows constructing an edit sequence for any pair (x, y). Tag prediction also improves explainability in the GEC task. For 3-tuple generation, a divided transformer decoder is used, and the tag and span predictions are located between its parts. {S}eq2{E}dits approach is similar to RedPenNet in the following (i) generation of spans and replacement tokens within the same autoregressive step, (ii) using Pointer Networks to predict spans. The difference between compared approaches is:
\begin{enumerate*}
\item RedPenNet uses a single decoder stack to generate tokens and spans. 
\item The {S}eq2{E}dits approach is different in terms of generating multi-token edits. According to~\citep{bryant-etal-2017-automatic}, an edit that has at least two tokens (multi-token edit) represents 10\
\item In the RedPenNet approach, decoder-specific positional encodings are added to the decoder inputs  at the bottom of the decoder stack. This allows the model to effectively utilize the order of multi-token edits. The approach presented in {S}eq2{E}dits does not clearly state the location in the divided Transformer decoder, where positional encodings can be utilized. The absence of such encodings can result in difficulty for the model in comprehending the order of the replacement tokens being inserted within the same span positions. 
\item RedPenNet uses a pre-calculated, task-specific version of the BPE decoder vocabulary to generate edit tokens, thus reducing the cost of the pre-softmax linear transformation.
\end{enumerate*}
 
 The Lasertagger~\citep{https://doi.org/10.48550/arxiv.1909.01187} approach deploys an autoregressive Transformer decoder to annotate the input sequence with tags from pre-calculated output vocabulary. With the limited size of the tags, vocabulary minimizes the cost of the pre-softmax linear transformation, making Lasertagger the fastest approach among the IOAR SeqToEdits architectures. However, the RedPenNet model presents several key differences: (i) it uses a BPE vocabulary instead of a tag vocabulary, (ii) it generates edits rather than tagging the input sequence, and (iii) it can produce a sequence of tokens for each edit.
 
 In the ESD\&ESC~\citep{chen-etal-2020-improving-efficiency} approach, the task of solving the GEC editing problem is divided into two subtasks: Erroneous Span Detection (ESD), where incorrect spans are identified through binary sequence tagging, and Erroneous Span Correction (ESC), where the correction of these spans is performed using a classic autoregressive approach that implies generation of edits for tokens surrounded by annotated span tokens. RedPenNet shares some similarities with this architecture, as it also utilizes autoregressive generation of a sequence of edit tokens, separated by control tokens that are part of the decoder vocabulary. The ESD\&ESC approach differs from RedPenNet in several key aspects.
 \begin{enumerate*}
 \item Firstly, RedPenNet predicts the span positions in a one-by-one manner at the decoder level, while the ESD\&ESC approach uses a separate encoder to generate spans. However, the ESD approach has the same limitations as the FF family, since the ESD tags may not always be consistent, leading to difficulties in maintaining consistency between interrelated edits. The ESC decoder during generation will not have the capability to fully rectify the situation, as it will be confined to the range of the annotated span tokens.
 \item RedPenNet approach is capable of decomposing neighboring errors in the input text into multiple edit operations, if necessary. Conversely, the ESD approach merges nearby errors in a single span. 
 \end{enumerate*}

 The Aggressive Decoding~\citep{https://doi.org/10.48550/arxiv.2106.04970} method accelerates the AR calculations for the task by utilizing the input tokens as drafted decoded tokens and autoregressively predicting only those portions that do not match. This leads to a significant improvement in inference speed. The disadvantage of the IGAD approach is that it requires the use of a shared vocabulary with the encoder during decoding. Therefore, even when the input and output sequences are the same, IGAD requires a significant number of floating point operations for the pre-softmax linear transformation in the decoder which is calculated using the formula: $O(v \cdot d \cdot l)$, where $v$ is the vocabulary size, $d$ is the model depth, and $l$ is the input length. This problem becomes more obvious in the case of using pre-trained multi-language models, which traditionally have larger encoder vocabularies and corresponding matrix embeddings. The impact of decoder vocabulary is analyzed in section~\ref{vocabularies}. Additionally, since the length of the output sequence in IGAD is directly tied to the length of the input sequence, the issue of quadratic complexity in attention mechanisms remains in the decoder. This can be a challenge when dealing with long sequences and requires the use of specialized transformer architectures in the decoder. 
 
 It is worth mentioning, that the Highlight and Decode Technique described in our previous study~\citep{didenko-shaptala-2019-multi}. Similar to the Erroneous Span Detection (ESD) component in the ESD\&ESC approach, a binary sequence tagging model was used to identify incorrect spans. Subsequently, a broadcast binary sequence mask was element-wise multiplied to a special “highlight” embedding. The result of this operation was added to the encoder output at the bottom of the decoder stack. This allowed the decoder to predict the replacement tokens only for the “highlighted” spans. However, as outlined in the mentioned article, this approach had a list of limitations.

\section{Experiments} \label{experiments}
\subsection{UNLP 2023 Shared Task}
The UNLP-2023 conference hosted the first Shared Task~\citep{syvokon-romanyshyn-2023-shared} in GEC for Ukrainian. One of the primary difficulties in addressing the GEC problem for the Ukrainian language lies in the scarcity of high-quality annotated training examples — a common issue for Non-English GEC. The Ukrainian language also poses an additional challenge due to its rich morphological structure and fusional nature.~\citep{DBLP:journals/corr/abs-2103-16997}.
The foundation of this Shared Task was established by Grammarly's efforts to develop a corpus that has been professionally annotated for GEC and fluency edits in the Ukrainian language, referred to as the UA-GEC corpus.
The Shared Task consists of two tracks: (i) GEC-only, which focuses on automatically identifying and correcting grammatical errors in written text, and (ii) GEC+Fluency, which encompasses corrections for grammar, spelling, punctuation, and fluency. Given that the RedPenNet architecture is capable of handling any type of editing, including rephrasing, reordering words, and sentence splitting, we decided to participate in the GEC+Fluency track.

\textbf{GEC+Fluency Baseline:} \label{baseline} Furthermore, the organizers offered a baseline model~\footnote{\url{https://huggingface.co/osyvokon/mbart50-large-ua-gec-baseline}} based on facebook/mbart-large-50. This model was trained for a NMT task with the objective of autoregressively generating correct text from erroneous input. The score of baseline can be found in table~\ref{results-table}.

\subsubsection{Data}
In the case of GEC tasks, data is typically stored in the \texttt{m2} format~\citep{dahlmeier-ng-2012-better}, where each instance consists of a source text and a list of edits required to transform it into the target text. To adapt the \texttt{m2} training examples for the RedPenNet architecture, we (i) deployed the pre-trained encoder tokenizer to tokenize the erroneous input text, (ii) used the decoder tokenizer~\ref{vocabularies} to tokenize the edits correction strings, and (iii) converted the span offsets from the word count separated by spaces to the corresponding BPE tokens (sub-words) offsets.

\textbf{UA-GEC:}
In the GEC+Fluency track of the Shared Task, the participants were given access to the gec-fluency public dataset~\footnote{\url{https://github.com/asivokon/unlp-2023-shared-task}}. The training data comprises 32,734 examples, where 15,161 contain at least one annotated error edit, while 17,573 are error-free. The evaluation dataset consists of 1,506 dev set and 1,350 test set instances. In this Shared Task, two annotators annotated all examples from the development set of the dataset and some examples from the training set. They also annotated all examples from the evolution dev sets, as well as some examples from the training data. For the Shared Task, all the data was tokenized using the stanza library~\footnote{\url{https://stanfordnlp.github.io/stanza/}}. To categorize the dataset edits by error types, we utilized a set of 20 tags. They included 14 grammar types and 6 fluency types.

\textbf{Synthetic Data:} \ 
Much of the research on the GEC problem shows that the use of pre-generated synthetic data reduces model training time and also improves overall quality. 
For the UNLP 2023 Shared Task, we generated over $160K$ Ukrainian erroneous data sentences based on error-free texts taken from data corpora presented on lang.org.ua~\footnote{\url{https://lang.org.ua/uk/corpora/}} website.
For our error generation approach, we utilized mbart-large-50 as a pre-train model, which we trained using the back translation method~\citep{xie-etal-2018-noising} on the training data from the UA-GEC dataset. Our task was to transduce the error-free input text sequence into the erroneous one. The synthetic data generation model was trained on a Google Colab Premium GPU instance for $8$ epochs with a batch size of $4$, a learning rate of $1\mathrm{e}{-5}$, and a maximum input and output length of $128$ tokens each. The performance of the RedPenNet architecture trained on this pre-training data is presented in Table~\ref{syntetical}.

\subsubsection{Decoder vocabulary for Ukrainian GEC} \label{ukr_gec_vocabularies}
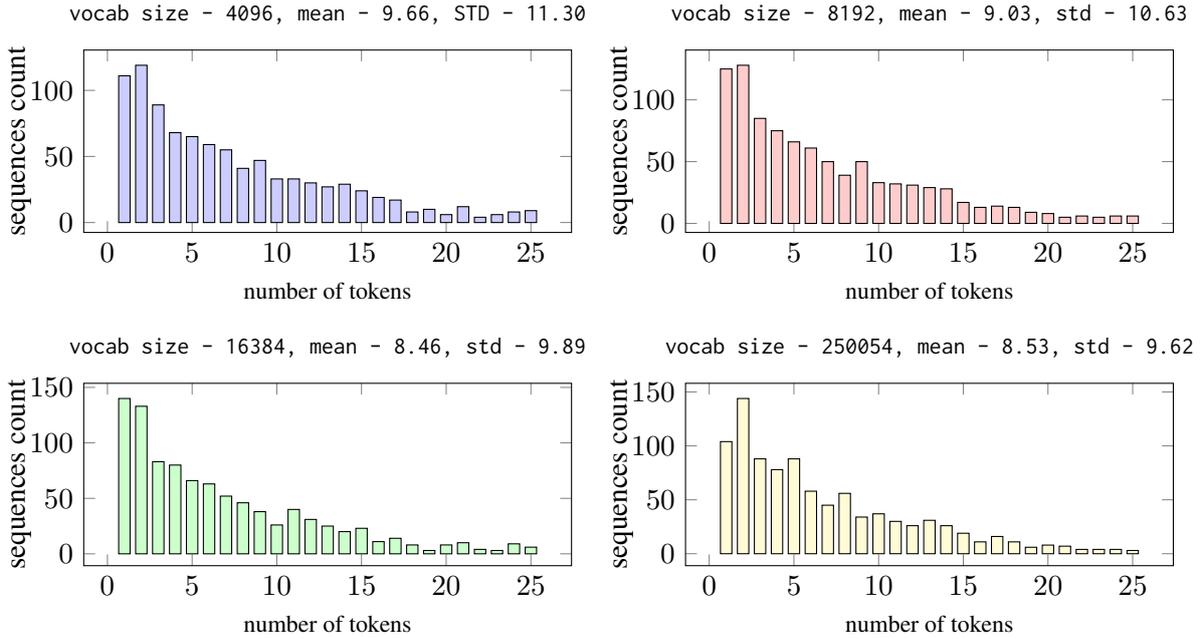
\begin{figure*}[ht]
\centering
\begin{tikzpicture}
\begin{groupplot}[group style={group size=2 by 2, vertical sep=2cm, horizontal sep=1.5cm}, height=4cm, width=8cm]
\nextgroupplot[title=\texttt{vocab size - 4096, mean - 9.66, STD - 11.30}, xlabel={number of tokens},x label style={font=\footnotesize\selectfont}, ylabel={sequences count}, y label style={below=1mm}, title style={font=\fontsize{9}{10}\selectfont}]
\addplot[ybar,bar width=0.15cm,height=0.1cm,fill=blue!20] coordinates {(1, 111) (2, 119) (3, 89) (4, 68) (5, 65) (6, 59) (7, 55) (8, 41) (9, 47) (10, 33) (11, 33) (12, 30) (13, 27) (14, 29) (15, 24) (16, 19) (17, 17) (18, 8) (19, 10) (20, 6) (21, 12) (22, 4) (23, 6) (24, 8) (25, 9)};
\nextgroupplot[title=\texttt{vocab size - 8192, mean - 9.03, std - 10.63}, xlabel={number of tokens},x label style={font=\footnotesize\selectfont}, ylabel={sequences count}, y label style={below=1mm}, title style={font=\fontsize{9}{10}\selectfont}]
\addplot[ybar,bar width=0.15cm,height=0.1cm,fill=red!20] coordinates {(1, 125) (2, 128) (3, 85) (4, 75) (5, 66) (6, 61) (7, 50) (8, 39) (9, 50) (10, 33) (11, 32) (12, 31) (13, 29) (14, 28) (15, 17) (16, 13) (17, 14) (18, 13) (19, 9) (20, 8) (21, 5) (22, 6) (23, 5) (24, 6) (25, 6)};
\nextgroupplot[title=\texttt{vocab size - 16384, mean - 8.46, std - 9.89}, xlabel={number of tokens},x label style={font=\footnotesize\selectfont}, ylabel={sequences count}, y label style={below=1mm}, title style={font=\fontsize{9}{10}\selectfont}]
\addplot[ybar,bar width=0.15cm,height=0.1cm,fill=green!20] coordinates {(1, 140) (2, 133) (3, 83) (4, 80) (5, 66) (6, 63) (7, 52) (8, 46) (9, 38) (10, 26) (11, 40) (12, 31) (13, 25) (14, 20) (15, 23) (16, 11) (17, 14) (18, 8) (19, 3) (20, 8) (21, 10) (22, 4) (23, 3) (24, 9) (25, 6)};
\nextgroupplot[title=\texttt{vocab size - 250054, mean - 8.53, std - 9.62}, xlabel={number of tokens},x label style={font=\footnotesize\selectfont}, ylabel={sequences count}, y label style={below=1mm}, title style={font=\fontsize{9}{10}\selectfont}]
\addplot[ybar,bar width=0.15cm,height=0.1cm,fill=yellow!20] coordinates {(1, 104) (2, 144) (3, 88) (4, 78) (5, 88) (6, 58) (7, 45) (8, 56) (9, 34) (10, 37) (11, 30) (12, 26) (13, 31) (14, 26) (15, 19) (16, 11) (17, 16) (18, 11) (19, 6) (20, 8) (21, 7) (22, 4) (23, 4) (24, 4) (25, 3)};
\end{groupplot}
\end{tikzpicture}

\caption{The x-axis depicts the number of tokens required to represent the concatenated correction of sequences for each \texttt{m2} instance. The y-axis represents the total count of the concatenated corrections extracted from the (\texttt{gec-fluency/valid.m2}) that meet a specific number of tokens. The results show that as the vocabulary size increases, the mean number of tokens needed to encode one concatenated correction decreases. However, when the vocabulary reaches 16,384, a vocabulary trained on corrections and frequent words outperforms the native vocabulary of mbart-large-50 in terms of the mean parameter.}
\label{fig:custom_decoder_tokenizers}
\end{figure*}

Multiple BPE decoder vocabularies were trained with varying sizes and evaluated based on the resulting output token count (refer to Figure~\ref{fig:custom_decoder_tokenizers}). A training text file was created specifically for this purpose, consisting of correction strings extracted from the \texttt{m2} edits. The (\texttt{gec-fluency/train.m2}) file from the UNLP-2023 Shared Task was used as the source. Also, we added additional 50,000 of the most frequent words from the Ukrainian Frequency dictionary of lexemes of artistic prose.\footnote{\url{http://ukrkniga.org.ua/ukr_rate/hproz_92k_lex_dict_orig.csv}} to the vocabularies training text file extracted from (\texttt{gec-fluency/train.m2}). The (\texttt{gec-fluency/valid.m2}) was utilized to evaluate and compare the different sizes of the decoder vocabularies.

The evaluation was performed by extracting and concatenating the correction strings from all edits for each annotated \texttt{m2} sentence into a space-separated sequence. This sequence was then tokenized using different decoder vocabularies.
Example:
annotated \texttt{m2} sentence:
\begin{quote}
\fontencoding{T2A}\selectfont  
\texttt{S Нечіткі бенефітс співпраці , натомість вихначені зобовязання}\\
\texttt{A 5 6|||Spelling|||визначені|||...|||0}\\
\texttt{A 6 7|||Spelling|||зобов'язання|||...|||0}\\
\texttt{A 1 2|||Spelling|||бенефіціари|||...|||1}\\
\texttt{A 5 6|||Spelling|||визначені|||...|||1}\\
\texttt{A 6 7|||Spelling|||зобов’язання|||...|||1}\\
\texttt{A 7 7|||Punctuation|||.|||...|||1}
\end{quote}

concatenated edits corrections:
\begin{quote}
\fontencoding{T2A}\selectfont 

визначені зобов'язання бенефіціари визначені зобов’язання .
\end{quote}

To compare the advantages of using a shorter task-specific decoder vocabulary for the Ukrainian GEC task, we will use the mbart-large-50 baseline model~\ref{baseline} as a reference. For this model, the number of operations required for the pre-softmax linear transformation is $(1024 \cdot 250,054)$ = 256,055,296 floating-point operations. In contrast, our trained vocabulary with a size of 16,384 performs the same task using only $(1024 \cdot 16,384)$ = 16,777,216 operations while maintaining a smaller encoding length than the baseline.

\begin{table*}[ht]
\footnotesize
\centering
\begin{tabular}{|l|c|ccc|ccc|}
\hline
\textbf{Approach} & {\textbf{min edit}} & \multicolumn{3}{c|}{\textbf{dev}}  & \multicolumn{3}{c|}{\textbf{test}} \\      \cline{3-8}
                       & \textbf{prob} & \textbf{P} &  \textbf{R}  & $\mathbf{F_{0.5}}$ & \textbf{P} &  \textbf{R}  & $\mathbf{F_{0.5}}$ \\ \hline
Hunspell                                   &   -   &   12.9    &  04.0   &   08.9   &   -   &  -   &   -   \\ 
Langtool(free)                             &   -   &   21.8   &  05.9   &   14.2   &   -   &  -   &   -   \\ 
Langtool(free)+Hunspell                    &   -   &   19.1   &  09.0   &   15.6   &   -   &  -   &   -   \\ 
MBart-50$_{\textsc{large}}$                &       &   67.51   &  39.48   &   59.11   &   73.06   &  44.36   &   64.69   \\ 
RPN(XLM$_{\textsc{base}}$)                 &   0.94 &   74.9    &  31.2   &   58.51   &   -   &  -   &   -   \\ 
RPN(R$_{\textsc{large}}$)                  &   0.95   &   75.31   &  35.11    &   \textbf{61.28}   &   76.54   &  41.93    &   \textbf{65.69}  \\  \hline
\specialcell{1$\times$RPN(XLM$_{\textsc{base}}$)\\ + 2$\times$RPN(R$_{\textsc{large}}$)}$\ast$  &  58.5/59   &   80.28   &  36.58    &    \textbf{64.80}   &   80.86   &  41.03    &   \textbf{67.71} \\  \hline
\end{tabular}
\caption{\label{results-table} displays the performance comparison between RedPenNet (RPN) and other existing public methods on the UA-GEC+Fluency dataset. The \textit{min edit prob} column shows the edit probability threshold required for accepting an edit.}
\end{table*}

\begin{table} \label{syntetical}
\footnotesize
\centering
\begin{tabular}{|l|c|c|c|}
\hline
\multicolumn{1}{|c|}{Approach} & \multicolumn{3}{c|}{\textbf{UA-GEC+Fluency (dev)}} \\ \cline{2-4} 
\multicolumn{1}{|c|}{} & \textbf{P}           & \textbf{R}           & $\mathbf{F_{0.5}}$          \\ \hline
\specialcell{Synt. pre-train \\ \& freeze encoder}   &  08.0   &  08.9    &   08.2 \\ \hline
Synt. pre-train  &   07.18   &  \textbf{17,27}   &   08.13 \\  
\hline
\end{tabular}
\caption{\label{syntetical} Performance of RPN(R$_{\textsc{large}}$) after pre-training on synthetic data.}
\end{table}

\subsubsection{Model Configuration}
\textbf{Encoders:} \
For the encoder part of RedPenNet, we chose pre-trained models from those available on the Hugging Face Hub~\footnote{\url{https://huggingface.co/models}}. Our main requirement during the selection process was that the models were trained on the Ukrainian language corpora. We built and compared a few models based on different encoders:  \textbf{RPN(R$_{\textsc{large}}$)} – RoBERTa Large~\footnote{\url{https://huggingface.co/benjamin/roberta-large-wechsel-ukrainian}} transferred to Ukrainian using the method from the NAACL2022 paper~\citep{minixhofer-etal-2022-wechsel}, \textbf{RPN(XLM$_{\textsc{base}}$)} – a smaller version of the XLM-RoBERTa~\footnote{\url{https://huggingface.co/ukr-models/xlm-roberta-base-uk}} model with only Ukrainian and some English embeddings left. Comparative results for these models can be seen in the table~\ref{results-table}

\textbf{Decoder:} \
We utilized a shallow RedPenNet Decoder stack~\ref{decoder} for the decoder part of our architecture. It consists of two layers, and we kept the model depth and dropout parameters the same as the encoders. We utilized a previously computed decoder vocabulary (refer to Section~\ref{ukr_gec_vocabularies}) which was set to a size of $16,384$.

\textbf{Setup:} \
Tensorflow 2 on a Google Colab TPU instance was used for training and evaluation. In most of the combinations, we conducted pre-training on synthetic data for $20$ epochs, followed by training on UA-GEC erroneous data for $30$ epochs using a batch size of $32$ and a learning rate of $2\mathrm{e}{-5}$. Afterward, we fine-tuned the model on UA-GEC (erroneous + error-free) data for $5$ epochs, with a batch size of $16$ and a learning rate of $5\mathrm{e}{-6}$. In the Results section~\ref{results}, we present the results of the approaches that showed the best performance.

\subsubsection{Evaluation} \label{evaluation}
For the evaluation, the organizers of the Shared Task provided the script based on Errant~\footnote{\url{https://github.com/chrisjbryant/errant}}. Although Errant isn’t able to handle specific error types in Ukrainian, it is common practice to use this library for other non-English languages, such as Spanish~\citep{davidson-etal-2020-developing-nlp}.
We have also evaluated scores on the free version of LanguageTool and Hunspell for comparison~\ref{results-table}.

In RedPenNet, we implemented a \textit{minimum edit probability} parameter to filter out low-probability edits and to improve precision at the cost of recall. To achieve this, we averaged the probabilities of all predicted edit tokens, as well as the predicted start span and end span for each edit. We assessed the probability of all edits in the model output and discarded those that have probabilities below the \textit{minimum edit probability}. All edits with probabilities surpassing the threshold were applied. A similar prediction filtering method for GEC was proposed in the GECToR paper and was called “Inference tweaking”. And in both cases, the method proved to be effective in improving precision.
We also experimented with an iterative correction process, where the output of a previous correction round is used as input for the next one.

\textbf{Ensembles:} \label{ensembles}
To create an ensemble of the RedPenNet models, we calculated the average edit probabilities and applied an algorithm that follows the subsequent scenario:
\begin{enumerate*}
    \item For matched edits, we summed their probabilities.
    \item For intersecting edits, we choose the more probable one.
    \item We kept all remaining non-intersecting edits in the result.
\end{enumerate*}
Then we tuned the \textit{minimum edit probability} parameter to maximize the F0.5 score on the UA-GEC+Fluency dev set.

\subsubsection{Results} \label{results}
We began by pre-training models solely on erroneous data, as proposed in the GECToR research. During this stage, we froze encoders and used synthetic data for pre-training. In the next stage, we unfroze the encoders and trained the models on UA-GEC erroneous data. Our experiments indicate that a low learning rate of $\pm 5\mathrm{e}{-6}$, a small batch size, a few training steps (less than epoch), and an increase in dropouts to $\pm0.2$ are useful during the initial stages of fine-tuning on a combination of (erroneous + error-free) data. This approach enables us to capture a good checkpoint when the model shifts from recall to precision.

To implement ensembles, we trained two RPN(R$_{\textsc{large}}$) models and one RPN(XLM$_{\textsc{base}}$) model. The only difference between the two RPN(R$_{\textsc{large}}$) models is that one of them was trained on erroneous data before being trained on (error-free + erroneous data). The model that was trained only on (error-free + erroneous data) has higher recall.

To enhance the quality of the results, we performed two rounds of iterative correction and applied the ensemble technique to the output of each round. During the first iteration, we set the \textit{minimum edit probability} to $0.585$, and for the second iteration, it was set to $0.59$. During iterative correction, we selected the value of the \textit{minimum edit probability} parameter that maximizes the precision score.

During the experiment, we demonstrated that our custom architecture, RedPenNet can be applied to the GEC task, with performance that competes with large Seq2Seq models like mbart-large-50 and significantly outperforms classical algorithmic approaches.

\subsection{BEA 2019 Shared Task}
To further demonstrate the capabilities of the RedPenNet architecture, we applied it to the BEA-2019 Shared Task on English GEC.

\textbf{Data:}
The combination of erroneous data obtained from several sources was used for pre-training. We used 20 million samples from the synthetic \textit{tagged corruption} dataset~\citep{stahlberg-kumar-2021-synthetic}\footnote{\url{https://huggingface.co/datasets/liweili/c4_200m}}, approximately 500K English samples from the ~\citep{rahman2022judge} study, and around 500K English samples from the \textit{lang-8} dataset. The data was sampled in the following proportions: 50\

\textbf{Model Configuration:}
We trained several different-sized RedPenNet models: two based on XLNet~\footnote{\url{https://huggingface.co/xlnet-large-cased}} pre-trains - RPN(XLN${\textsc{base}}$) and RPN(XLN${\textsc{large}}$), and two models based on Muppet Roberta~\footnote{\url{https://huggingface.co/facebook/muppet-roberta-large}}: RPN(MPR${\textsc{base}}$) and RPN(MPR${\textsc{large}}$).

The decoder stack consists of two layers, and we utilized a pre-computed decoder vocabulary trained on text corrections extracted from \texttt{ABC.train.gold.bea19.m2}. The chosen vocabulary size is $8192$.

For pre-training, we conducted 500K steps with a batch size of $128$ for $\textsc{base}$ models and $64$ for $\textsc{large}$, setting the learning rate to $3\mathrm{e}{-5}$. For fine-tuning, we performed 4-6 epochs (depending on the model) to obtain the maximum $F_{0.5}$ score on the \textit{W\&I+LOCNESS} dev set. During fine-tuning, we used a batch size of $32$ and a learning rate of $5\mathrm{e}{-6}$ for all models.

\begin{table}[ht]
\small
\centering
\caption{BEA-2019 (Test)}
\label{table:bea-2019-results}
\begin{tabular}{lccc}
\hline
\textbf{Model}                   & \textbf{P} & \textbf{R} & \textbf{$F_{0.5}$} \\ \hline
~\citep{qorib-etal-2022-frustratingly}$\ast$                      & 86.6       & 60.9       & 79.9               \\ \hline
~\citep{lichtarge-etal-2020-data}       & 75.4       & 64.7       & 73.0               \\
~\citep{https://doi.org/10.48550/arxiv.2005.12592}       & 79.4       & 57.2       & 73.7               \\  
~\citep{stahlberg-kumar-2021-synthetic}      & 77.7       & 65.4       & 74.9               \\
~\citep{rothe-etal-2021-simple}             & -          & -          & 75.9               \\ \hline
RPN(MPR$_{\textsc{base}}$)                      & 80,80       & 56,71       & 74,47               \\
4$\times$RPN ensemble & \textbf{86.62}       & 54.80       & \textbf{77.60}               \\\hline
\end{tabular}
\normalsize
\caption{\label{BEA-2019 results} A comparison of the performance of various modern GEC approaches, including RedPenNet on the BEA-2019 test set.~\citep{qorib-etal-2022-frustratingly}$\ast$ provides results of combination several systems outputs. }
\end{table}

\textbf{Evaluation and Results:}
We evaluated RedPenNet models on \textit{W\&I+LOCNESS} test set. For our best result, we used an ensemble of RPN(XLN${\textsc{base}}$), RPN(XLN${\textsc{large}}$), RPN(MPR${\textsc{base}}$) and RPN(MPR${\textsc{large}}$) models. We merged the output using the same scenario as for the UNLP 2023 Shared Task~\ref{ensembles} and determined the best \textit{minimum edit probability} to be $0.68$. Interestingly, the second round of processing, in which the outputs from the previous round served as model inputs, did not lead to an improvement in the $F_{0.5}$ score. As it is shown in Table~\ref{BEA-2019 results}, our approach yields state-of-the-art results on BEA-2019 (Test) benchmark, surpassed only by the System Combination result by~\citep{qorib-etal-2022-frustratingly}. Furthermore, it is worth mentioning that the RedPenNet ensemble consisting of four ${\textsc{base}}$/${\textsc{large}}$ models outperforms the BEA-2019 (Test) $F_{0.5}$ score of the T5-XXL 11B model from~\citep{rothe-etal-2021-simple} study.

\section{Conclusion}
While there has been a significant amount of research in the field and many tailored architectures have been proposed, a universally accepted neural architecture for text editing tasks that involves highly similar input and output sequences has yet to be established. This has prevented the creation of an industry standard that can be included in default toolkits for popular machine learning libraries and MLOps tools. Our proposed RedPenNet is an attempt to create a universal neural architecture that is not overloaded with design nuances and is capable of implementing any source-to-target transformation using a minimal number of autoregressive steps. The RedPenNet architecture is a classic transformer, and the only differences lie in how we form decoder input embeddings and interpret outputs and attention scores.

\section*{Limitations}
While the RedPenNet approach has demonstrated several strengths, such as superior inference capabilities for seq2seq tasks with highly similar inputs and outputs, and some advantages over other SeqToEdits approaches highlighted in the Related Works~\ref{related} section, it is not without its limitations:

Due to the tailored architecture of RedPenNet, there are no off-the-shelf solutions for data preprocessing, training, and fine-tuning, as is the case of tasks such as common classification or sequence-to-sequence. Consequently, it is not possible to use convenient tools like the HuggingFace Estimator or cloud platforms for rapid model fine-tuning and deployment.

Additionally, the implementation of a non-greedy beam search approach is complicated by the presence of multiple sequence outputs.

One more fundamental limitation is that for each edit, the model needs to generate at least two tokens (SEP, token). This does not provide an advantage in reducing the number of autoregressive steps, particularly for short and error-crowded sentences.

Additionally, while RedPenNet has the ability to express any type of input sequence transformation through a number of editing operations, it may not be able to express a single “conceptual” edit, such as transferring a word within a sentence, using a single edit operation. In such cases, two edits — deletion and insertion — may be required to accomplish the desired transformation.

\section*{Ethics Statement}
Our study focuses on the development of a neural architecture for text editing tasks. The research was conducted in accordance with ethical principles, and no sensitive or personal data was used or collected during the study. The UA-GEC dataset and corpora presented on lang.org.ua used in the study have been obtained from public sources, and their authors assure the privacy and confidentiality of the original texts. The results of the study are intended to improve the efficiency and accuracy of text writing and may be useful for other NLP tasks. We ensure that the study does not raise any ethical concerns or has no negative impact on individuals or groups.

\section*{Acknowledgments}
We would like to acknowledge and give our thanks to WebSpellChecker LLC for the support and resources allocated to this project. We are also grateful to the WebSpellChecker team, especially Julia Shaptala and Viktoriia Biliaieva, for their assistance and advice during the competition. We are expressing our gratitude to the Program Committee reviewers for organizing the first Shared Task in Grammatical Error Correction (GEC) for Ukrainian, their guidance and insightful recommendations.

\bibliography{anthology,custom}
\bibliographystyle{acl_natbib}

\end{document}